%% file: main.tex
\documentclass[runningheads]{llncs}
\usepackage{graphicx}
\usepackage{comment}
\usepackage{amsmath,amssymb} 
\usepackage{color}
\usepackage{symbols}
\usepackage{enumitem}
\usepackage{ctable} 

\definecolor{citecolor}{RGB}{119,185,0}
\definecolor{gold}{RGB}{218,165,32}

\usepackage[pagebackref=true,breaklinks=true,letterpaper=true,colorlinks,citecolor=citecolor,bookmarks=false]{hyperref}

\begin{document}
\pagestyle{headings}
\mainmatter
\def\ECCVSubNumber{3205}  

\title{UFO$^2$: A Unified Framework towards Omni-supervised Object Detection} 
\titlerunning{UFO$^2$: A Unified Framework towards Omni-supervised Object Detection}
\author{Zhongzheng Ren\inst{1,2\footnotemark[1]} \and Zhiding Yu\inst{2} \and
Xiaodong Yang\inst{2\footnotemark[1]} \and Ming-Yu Liu\inst{2} \and
 \\ Alexander G. Schwing\inst{1} \and Jan Kautz\inst{2} } 
\authorrunning{Z. Ren et al.}
\institute{University of Illinois at Urbana-Champaign \and NVIDIA 
}

\maketitle
\footnotetext[1]{Work partially done at NVIDIA.}
\input{sections/abstract.tex}
\input{sections/intro.tex}
\input{sections/related.tex}

\input{sections/approach.tex}

\input{sections/label.tex}
\input{sections/experiments.tex}
\input{sections/conclusion.tex}

\noindent\textbf{Acknowledgement:} ZR is supported by Yunni \& Maxine Pao Memorial Fellowship. This work is supported in part by  NSF under Grant No.\ 1718221 and MRI \#1725729, UIUC, Samsung, 3M, Cisco Systems Inc.\ (Gift Award CG 1377144) and Adobe.

\clearpage
\bibliographystyle{splncs04}
\bibliography{reference}

\clearpage
\input{sections/supp.tex}

\end{document}

%% file: sections/abstract.tex

\begin{abstract}
Existing work on object detection often relies on a single form of annotation: the model is trained using either accurate yet costly bounding boxes or cheaper but less expressive image-level tags. However, real-world annotations are often diverse in form, which challenges these existing works. In this paper, we present UFO$^2$, a unified object detection framework that can handle different forms of supervision simultaneously. Specifically, UFO$^2$ incorporates strong supervision (\eg, boxes), various forms of partial supervision (\eg, class tags, points, and scribbles), and unlabeled data. Through rigorous evaluations, we demonstrate that each form of label can be utilized to either train a model from scratch or to further improve a pre-trained model. We also use UFO$^2$ to investigate budget-aware omni-supervised learning, \ie,  various annotation policies are studied under a fixed annotation budget: we show that competitive performance needs no strong labels for all data. Finally, we demonstrate the generalization of UFO$^2$,  detecting more than 1,000 different objects without bounding box annotations.

\keywords{Omni-supervised, Weakly-supervised, Object Detection.}
\end{abstract}

%% file: sections/intro.tex
\section{Introduction}

State-of-the-art object detection methods benefit greatly from supervised data, which comes in the form of bounding boxes on many datasets. However, annotating images with bounding boxes is time-consuming and hence expensive.  To ease this dependence on expensive annotations, `\textit{omni-supervised learning}'~\cite{RadosavovicDGGH18} has been proposed: models should be trained via all types of available labeled data plus internet-scale sources of unlabeled data. 

\begin{figure}[t]
\centering
\includegraphics[width=0.9\textwidth]{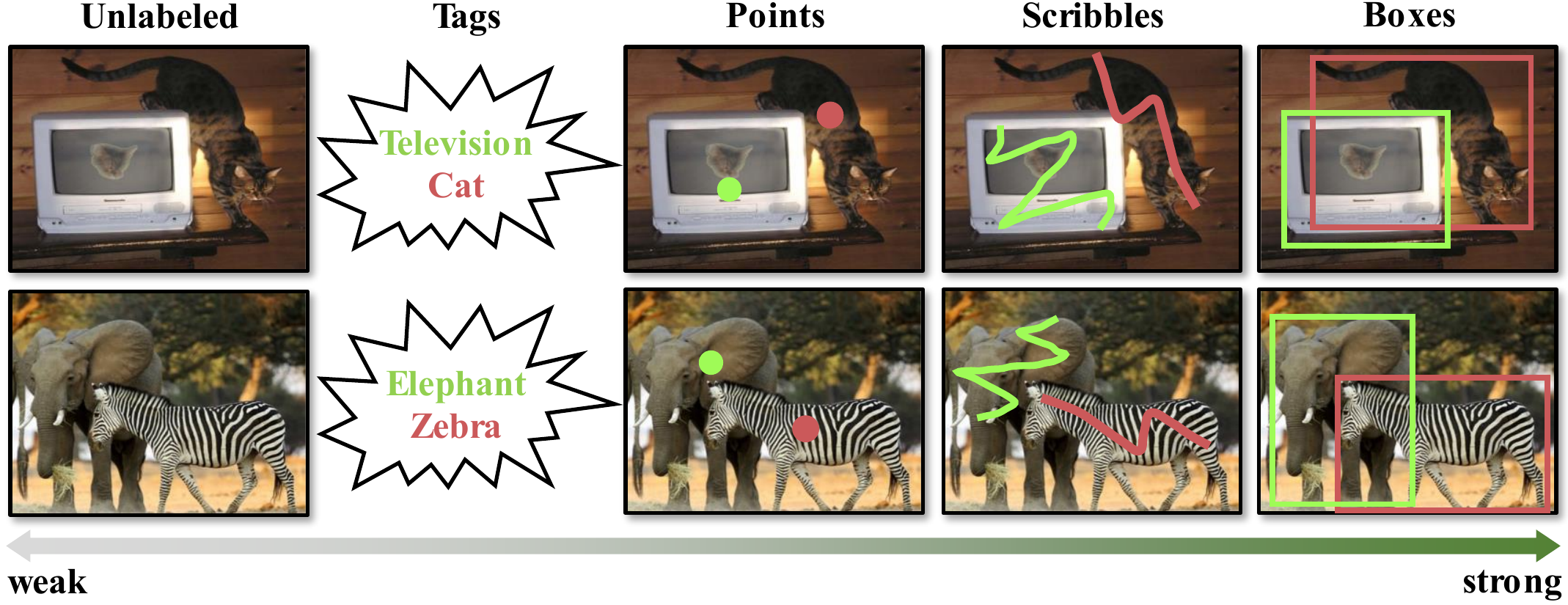} 
\caption{Illustrative example of the supervision hierarchy.}
\label{fig:teaser}
\end{figure}

Omni-supervised learning is particularly beneficial in practice. Compared to the enormous amounts of visual data uploaded to the internet (\eg, over 100 million photos uploaded to Instagram every day~\cite{instgram}; 300 hours of new video on YouTube each minute~\cite{youtube}), fully-annotated training data remains a negligible fraction. Most data is either  unlabeled, or comes with a diverse set of weak labels. Hence, directly leveraging web data often requires handling labels that are incomplete, inexact, or even incorrect (noisy).

Towards the goal of handling real-world messy data, we aim to study omni-supervised \emph{object detection} where a plethora of unlabeled, partially labeled (with image-level class tags, points, or scribbles), and strongly labeled (with bounding boxes) images are utilized to  train detection models. Examples of the considered supervisions are shown in Fig.~\ref{fig:teaser}. Designing a framework for omni-supervised detection is non-trivial. A big challenge is the conflict of the different architectures that have been proposed for each annotation. To address this issue, prior work either ensembles different networks trained from different annotations~\cite{khodabandeh2019robust,UijlingsPF18-Revisiting} or uses iterative knowledge distillation~\cite{Budget,krishna-cvpr2016}. However, the conflict between different modules remains as it is largely addressed in a post-processing step. 

In contrast, we propose UFO$^2$, a unified omni-supervised object detection framework that addresses the above challenges with a principled and computationally efficient solution. To the best of our knowledge, the proposed framework is the first to simultaneously handle direct supervision, various forms of partial supervision, and unlabeled data for object detection. UFO$^2$ (1) integrates a \textbf{unified task head} which handles various forms of supervision (Sec.~\ref{sec:head}), and (2) incorporates a \textbf{proposal refinement} module that utilizes the localization information contained in the labels to restrict the assignment of object candidates to class labels (Sec.~\ref{sec:roi-refine}). Importantly, the model is \textbf{end-to-end trainable}. 

We note that assessing the efficacy of the proposed approach is non-trivial. Partial labels are hardly available in popular object detection data~\cite{mscoco}. We thus create a simulated set of partial annotations, whose labels are synthesized to closely mimic human annotator behavior (Sec.~\ref{sec:label}). We then conduct rigorous evaluations  to show that: (1) each type of label can be effectively utilized to either train a model from scratch or to boost the performance of an existing model (Sec.~\ref{exp:from-0}); (2) a model trained on a small portion of strongly labeled data combined with other weaker supervision  can perform comparably to a fully-supervised model under a fixed annotation budget, suggesting a better annotation strategy in practice (Sec.~\ref{exp:pizza}); (3) the proposed model can be seamlessly generalized to utilize large-scale classification (only tags are used) data. This permits to scale the detection model to more than 1,000 categories (Appendix~\ref{exp:extension}).

%% file: sections/related.tex

\section{Related Work}
In the following we first discuss related works for each single supervision type. Afterwards we introduce prior works to jointly leverage multiple labels for visual tasks. Training data usage of prior object detection works are given in Tab.~\ref{table:related}. 

\begin{table*}[t]
\centering
\begin{tabular}{c | c | c | c | c | c | c}
\specialrule{.15em}{.05em}{.05em} 
B & T & P & S & B+U & B+T & B+T+P+S+U \\
\hline
\cite{dpm,rcnn,ren16faster,yolo,ssd,Law_2018_ECCV} & 
\cite{Bilen16,tang2017multiple,Zhang_2018_CVPR,ren-cvpr2020} &
\cite{extreme-click,train-click} &
None &
\cite{RadosavovicDGGH18,Rosenberg,Chen0SDG18} &
\cite{InoueFYA18,RedmonF17,Gao_2019_ICCV,Yang_2019_CVPR,UijlingsPF18-Revisiting} & 
\textbf{UFO$^2$(ours)}\\
\specialrule{.15em}{.05em}{.05em} 
\end{tabular}
\caption{Summary of related works for object detection using different labels. (B: boxes, T: tags, P: points, S: scribbles, U: unlabeled.) }
\label{table:related}
\end{table*}

\paragraph{\textbf{Object Detection.} }
Object detection has been one of the most fundamental problems in computer vision research. Early works~\cite{dpm} focus on designing hand-crafted features and multi-stage pipelines to solve the problem. Recently, Deep Neural Nets (DNNs) have greatly improved the performance and  simplified the frameworks. Girshick~\etal~\cite{fastrcnn,rcnn}  leverage DNNs to classify and refine pre-computed object proposals. However, those methods are slow during inference because the proposals need to be computed online using time-consuming classical  methods~\cite{ss,eb}. To alleviate this issue, researchers have designed  DNNs that learn to generate proposals~\cite{ren16faster,he2017maskrcnn} or one-shot object detectors~\cite{ssd,yolo}. Recently, top-down solutions have emerged,  re-formulating detection as key-point estimation~\cite{Law_2018_ECCV}. 
These methods achieve impressive results. However, to train these methods, supervision in the form of accurate localization information (bounding boxes) for each object is required. Collecting this supervision is not only costly in terms of time and money, but also prevents detectors from generalizing to new environments with scarce labeled data. 

\paragraph{\textbf{Weakly-supervised Learning.} }
Weak labels in the form of image-level category tags are studied in various tasks~\cite{khoreva_CVPR17,zhou2015cnnlocalization,singh-2017,PapandreouCMY15}. For object detection, existing works~\cite{Bilen16,tang2017multiple,Zeng_2019_ICCV,ren-cvpr2020} formulate a multiple instance learning task:  the input image acts as a bag of pre-computed proposals~\cite{ss,mcg,eb} and several most representative proposals are picked as detections. Bilen and Vedaldi~\cite{Bilen16} are among the first to implement the above idea in an end-to-end trainable DNN. Follow-up works boost the performance by including extra information, such as spatial relations~\cite{Peng_2015_ICCV,Zhang_2018_CVPR,ren-cvpr2020} or context information~\cite{KantorovOCL16,ren-cvpr2020}. In addition, better optimization strategies like curriculum learning~\cite{zigzag}, self-taught learning~\cite{JieWJFL17}, and iterative refinement~\cite{Peng_2015_ICCV,Shen_2018_CVPR,Ge_2018_CVPR} have shown success. However, due to the limited representation ability of weak labels, these methods often suffer from two issues: (1) they cannot differentiate multiple instances of the same class when  instances are spatially close; (2) they tend to focus on the most discriminative parts of an object instead of its full extent.  This suggests that training object detectors solely from weak labels is not satisfactory and motivates  to study a  hybrid approach. 

\paragraph{\textbf{Partially-supervised Learning.} }
Points and scribbles are two user-friendly ways of interacting with machines. Thus they are  widely used in various visual tasks such as semantic segmentation~\cite{XuCVPR2015,LinDJHS16ScribbleSup,bearman_point}, instance segmentation~\cite{zhou2019bottomup},  and image synthesis~\cite{park2019SPADE}. From a  data annotation perspective, these partial labels are  easier to acquire than labeling bounding boxes or masks~\cite{LinDJHS16ScribbleSup}. However, partial labels are in general understudied in object detection. A few examples on this topic include Papadopoulos \etal~\cite{extreme-click,train-click} which collect click annotation for the VOC~\cite{pascal} dataset and train an object detector through iterative multiple instance learning. Different from their approach, however, we propose an end-to-end trainable framework and evaluate on more challenging data~\cite{mscoco,lvis}.

\paragraph{\textbf{Semi-supervised Learning.} }
Semi-supervised learning~\cite{ssl,oliver2019benchmark,ren-ssl2020} aims to augment the limited annotated training set with large-scale unlabeled data to boost the performance. Recent approaches~\cite{MixMatch,vat,Tarvainen2017meanteacher,xie2019uda,Zou_2018_ECCV,zou2019confidence,ren-ssl2020} on classification often utilize unlabeled data through self-training combined with various regularization techniques including consistency regularization through data augmentation~\cite{MixMatch,Tarvainen2017meanteacher,xie2019uda}, entropy minimization~\cite{vat,Lee2013pseudo}, and weight decay~\cite{MixMatch}. In this paper, we adopt the entropy regularization~\cite{vat} and pseudo-labeling~\cite{Lee2013pseudo} methods to efficiently utilize unlabeled data.

For object detection, Rosenberg \etal~\cite{Rosenberg} demonstrate that self-training is still effective. Ensemble methods~\cite{RadosavovicDGGH18,MixMatch} and  representation learning~\cite{Chen0SDG18,he2019moco,doersch2015unsupervised,ren-cvpr2018} are shown to be useful. Nevertheless, these methods are heavily pipelined and usually assume existence of a portion of strong labels  to initialize the teacher model. In contrast, our UFO$^2$ learns from an arbitrary combination of either strong or partial labels and unlabeled data, it is unified and end-to-end trainable. 

\paragraph{\textbf{Omni-supervised Learning.} }
Omni-supervised learning is a more general case of semi-supervised learning in the sense that several types of available labels are mixed to train visual models jointly. Xu~\etal~\cite{XuCVPR2015} develop a non-deep learning method to jointly utilize image tags, partial supervision, and unlabeled data for semantic segmentation and perform competitively. Ch{\'e}ron~\etal~\cite{Cheron} extend this idea to video data by training an action localization network using various labels. However, their method cannot deal with unlabeled data. 

For object detection, prior works~\cite{RedmonF17,InoueFYA18,Gao_2019_ICCV,Yang_2019_CVPR,UijlingsPF18-Revisiting} have studied to combine bounding boxes and image tags. However, these methods are either pipelined and iterative~\cite{Gao_2019_ICCV,InoueFYA18,UijlingsPF18-Revisiting} or require extra activity labels and human bounding boxes to guide the detection~\cite{Yang_2019_CVPR}. Compared to those works, the proposed framework can handle more types of labels and, importantly, our proposed approach is end-to-end trainable.

%% file: sections/approach.tex
\section{UFO$^2$}

We aim to solve omni-supervised object detection: a single object detector is learned jointly from various forms of labeled and unlabeled data. Formally, the training dataset contains two parts: an unlabeled set $\mathcal{U} = (u_i;i\in \{1,\ldots,|\mathcal{U}|\})$ and a labeled set $\mathcal{X}=(x_i; i\in \{1,\ldots,|\mathcal{X}|\})$. Each $x_i$ is associated with one annotated label coming in one of the following four forms: (1) accurate bounding boxes, (2) a single point on the object, (3) a  scribble overlaying the object in some form, or (4) image-level class tags. Note, for the first three forms of annotations  we also know the semantic class. In this paper, we make \textbf{no assumptions on labels}: every form of label can make up any fraction of the training data. This is in contrast to most prior work  on mixed supervision~\cite{Gao_2019_ICCV,UijlingsPF18-Revisiting,RedmonF17,Yang_2019_CVPR} which assumes a certain amount of strongly labeled data (bounding boxes) is always available.

\begin{figure*}[t]
\centering
\includegraphics[width=0.99\textwidth]{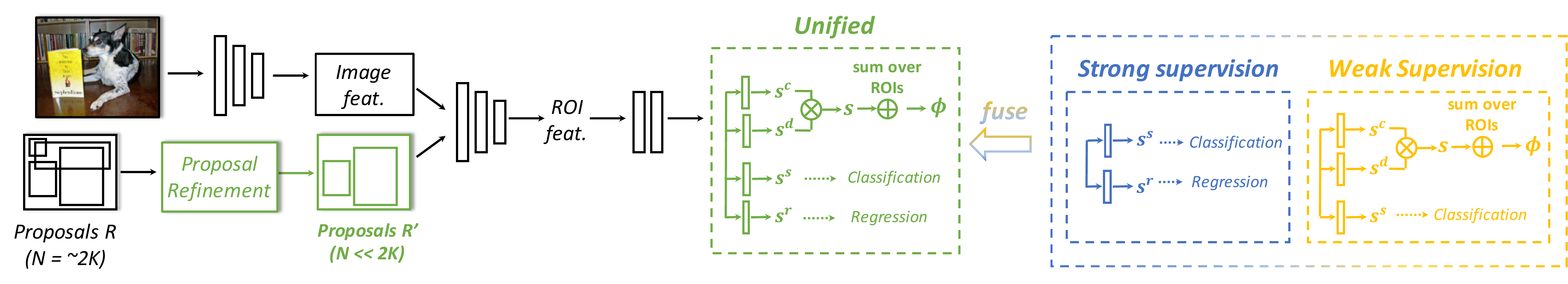} 
\caption{The UFO$^2$ framework: green modules are newly proposed in this paper.}
\label{fig:archi}
\end{figure*}

Since each form of annotation  has been separately studied in prior work, different frameworks have been specifically tailored for each annotation. In contrast, we present a novel unified framework UFO$^2$ which inherits merits of prior single supervision methods and permits to exploit arbitrary combinations of labeled and unlabeled data as shown in Fig.~\ref{fig:archi}. We introduce the specific solution to handle each supervision in Sec.~\ref{sec:tag}. We further devise an improved proposal refinement module~\cite{fastrcnn,ren16faster} so as to incorporate localization information in partial labels (see Sec.~\ref{sec:roi-refine}). 

\subsection{Unified Model}
\label{sec:head}
As shown in Fig.~\ref{fig:archi}, for a labeled input image $x\in\mathcal{X}$ or an unlabeled $u\in\mathcal{U}$, convolutional layers from an ImageNet pre-trained neural network are used to extract  image features. A set of pre-computed object proposals $R$ is refined to the set $R'$ and then used to generate ROI features through ROI-Pooling~\cite{he2017maskrcnn}. Note that not all the proposals are used since they are usually redundant. We discuss our  refinement technique in Sec.~\ref{sec:roi-refine}.  In our proposed model, the ROI features are processed via  several intermediate layers followed by a new task head as shown in Fig.~\ref{fig:archi} ({\color{citecolor} center, green}), which differs from classical methods. 

\subsubsection{Classical Methods.}
In strongly supervised frameworks~\cite{fastrcnn,ren16faster}, the task head consists of two fully-connected layers to produce the classification logits $s^s(r,c)\in\mathbb{R}$ for every region $r\in R'$ and class $c\in C$, and the region coordinates $s^r(r)\in \mathbb{R}^4$ for every region $r\in R'$ for bounding box regression. This is highlighted via a {\color{blue} blue box} in Fig.~\ref{fig:archi}.

In weakly-supervised frameworks~\cite{Bilen16,tang2017multiple,ren-cvpr2020} which handle image-level tags, the task head contains three fully-connected layers to produce a class confidence score $s^c(r,c)\in \mathbb{R}$, an objectness score $s^d(r,c)\in \mathbb{R}$, and similarly, classification logits $s^s(r,c) \in \mathbb{R}$ for every region $r\in R'$ and class $c\in C$ (Fig.~\ref{fig:archi} {\color{gold} yellow box}). 
The class confidence score $s^c(r,c)$ and objectness score $s^d(r,c)$ are first normalized via:
\begin{equation}
s^c(r,c) = \frac{\exp s^c(r,c)}{\sum_{c\in C}\exp s^c(r,c)},~\text{and}~
s^d(r,c) = \frac{\exp s^d(r,c)}{\sum_{r\in R}\exp s^d(r,c)}.
\label{eq:softmax}
\end{equation} 
They are then used for image-level classification. Also, $s^s(r,c)$ is used similarly for region classification using online-computed pseudo-labels. 

\subsubsection{UFO$^2$ Loss.} 
We propose to fuse both heads 
into a unified task head to produce the four aforementioned scores simultaneously as shown in Fig.~\ref{fig:archi} ({\color{citecolor} center green box}). A joint objective is optimized via
\begin{equation}
\mathcal{L_{\text{joint}}} \!=\! \mathcal{L}_{I} +  \frac{1}{|R'|} \!\sum_{r \in R', c \in C}\! \mathcal{L}_{R}(s^r(r), t(r)) + \mathcal{L}_{C}(s^s(r,c), y(r,c)) ,
\label{eq:total}
\end{equation} 
where $\mathcal{L}_{I}$ subsumes different losses for different labels and $\mathcal{L}_{C}, \mathcal{L}_{R}$ are standard cross-entropy loss and smooth-L1 loss for region classification and regression respectively. Moreover, $y(r,c)\in\{0,1\}$ and $ t(r)\in\mathbb{R}^4$ are either ground-truth region labels and regression targets from strong labels, or pseudo labels and pseudo targets generated online for partial labels and unlabeled data. We provide detailed explanations for $\mathcal{L}_I$ and how to generate pseudo labels $y(r,c)$ and pseudo targets $t(r)$ in the following. We discuss each form of annotation separately. 

\begin{figure*}[t]
\centering
\includegraphics[width=0.95\textwidth,trim=0 0 0 16pt]{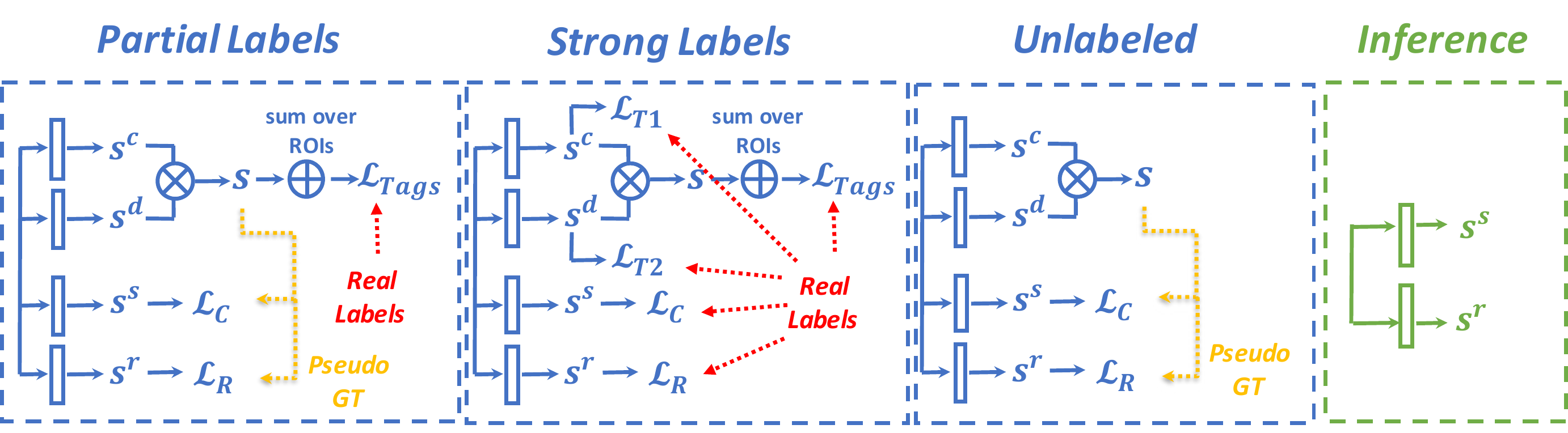} 
\caption{Task head behavior for training (w/ partial, strong, or no labels), and inference.}
\label{fig:heads}
\end{figure*}

\subsubsection{Tags.} 
\label{sec:tag}
As illustrated in Fig.~\ref{fig:heads} left,
when input images $x$ come with image-level class tags $q(c)\in\{0,1\}$ for class $c\in C$, we neither know the exact assignment of class labels to each proposal nor the exact target location. Therefore, we first compute the image scores via $s(r,c)=s^c(r,c)\cdot s^d(r,c)$, \ie, as a product of the class confidence score $s^c$ and the objectness score  $s^d$. Then image level evidence $\phi$ is obtained by summing up $s(r,c)$ across all regions:  $\phi(c)=\sum_{r\in R'}s(r, c)$. We then compute $\mathcal{L}_{\text{Tags}}$ as an image-level binary cross-entropy loss for multi-label classification: 
$$ 
\mathcal{L}_{\text{Tags}}(\phi, q) = -\sum_{c\in C}  q(c) \log \phi(c).
$$ 

For samples with image-level tags we set  $\mathcal{L}_{I} = \mathcal{L}_{\text{Tags}} $ in Eq.~\eqref{eq:total} during training. This yields semantically meaningful ROI scores $s(c,r)$, which can then be used to generate pseudo ROI-level ground-truth to augment the training via the two region-level losses $\mathcal{L}_C$ and $\mathcal{L}_R$ as detailed in Eq.~\eqref{eq:total}. We follow Ren \etal~\cite{ren-cvpr2020} to generate pseudo ground-truth, taking one or few diverse confident predictions. 

\subsubsection{Points \& Scribbles.}
\label{sec:ps}
Similar to image-level tags, points and scribbles also don't contain the exact region-level ground-truth. However, they provide some level of localization information (\eg, scribbles can be very rough or accurate depending on the annotator). Therefore, we employ the same loss developed for `Tags,' (illustrated in Fig.~\ref{fig:heads} left) but introduce extra constraints to restrict the assignment of ROIs to class labels based on the labels. Specifically, pseudo label $y(r,c)=1$ if and only if region $r$ contains the given point or scribble, and class $c$ is the same as the category label of this point or scribble. These constraints  filter out a lot of false-positives during training and help the framework  select high quality candidate regions. 

\subsubsection{Boxes.} 
\label{sec:box}
When the input image is annotated with bounding boxes, the most na\"ive solution is to directly train the network using $\mathcal{L}_{C}$ and $ \mathcal{L}_{R}$ losses: the real label and target are given and the scores $s^s$ and $s^r$ will be used for inference. Most supervised work~\cite{fastrcnn,ren16faster} follows the above procedure and impressive results are  achieved. Importantly, only applying these two losses in our  framework will not optimize the scores $s^c$ and $s^d$  when learning from strong labels. However, these two scores are  used as a `teacher' to compute pseudo ground-truth for optimizing $s^s$ and $s^r$ when partial labels are given, as described in the previous two sections. Hence, when training with mixed annotations, we found the `student' to  be stronger than the `teacher,' rendering weakly labeled data useless.

To address this concern, \ie, to enable training with mixed annotations, we found a \textit{balanced teacher-student model} to be crucial. Specifically, for any fully labeled sample we introduce three extra losses on the latent modules, \ie, on $s^c, s^d, \phi$,  as shown in Fig.~\ref{fig:heads} second column: 
\begin{equation}
\mathcal{L}_I= \mathcal{L}_{\text{Tags}}(\phi, q) + \frac{1}{|R|}\sum_{r \in R}(\mathcal{L}_{T1} (s^c, y, r)  + \mathcal{L}_{T2} (s^d, \psi, r) ).
\label{eq: strong}
\end{equation}
These three losses  provide a signal to  the `teacher' when using strong labels.
Specifically, since $s^c$ is normalized across all classes via a softmax, as mentioned in Eq.~\eqref{eq:softmax}, we can naturally apply as the first strong-teacher loss a standard cross-entropy on $s^c$ for region classification:
$$
\mathcal{L}_{T1} (s^c, y, r) = - \sum_{c\in C}  y'(c,r) \log s^c(c,r).
$$
Hereby $y'(c,r)=1$ for all  regions $r$ which overlap with any ground-truth boxes in class $c$ by more than a threshold. In practice, we set this threshold to $0.5$ and we use the class of the biggest overlapping ground-truth as the label if assignment conflicts occur. 
The second  strong-teacher loss encourages the latent distribution $s^d$ to approach the real objectness distribution. Hence we use a KL-divergence applied on $s^d$:
$$
\mathcal{L}_{T2} (s^d, \psi, r) =  \sum_{c\in C}   \psi(c,r) \log \frac{\psi(c,r) }{s^d(c,r)}.
$$
Here, $\psi(c,r)$ is constructed to represent the objectness of each ROI.  $\psi$ is zero initialized and $\psi(c,r)=\text{IoU}(r, r')$ for ground-truth region $r'$ with class $c$.  We then normalize $\psi$ across all $r \in R'$, following $s^d(c,r)$  normalization in Eq.~\eqref{eq:softmax}. 

In addition, we also construct an image-level class label $q$ from the ground-truth annotations and compute the  image-level classification loss $\mathcal{L}_{\text{Tags}}$ following the `Tags' setting. This loss term  improves  network consistency  when switching between partial labels and strong labels. 

\subsubsection{Unlabeled.} 
\label{sec:un}
For unlabeled data, we employ a simple yet effective strategy as shown in Fig.~\ref{fig:heads} third column. We use a single threshold $\tau$ on $\phi(c)$ to first pick out a set of confident classes $\hat{q}(c)$. This  set of classes is used as tags to train the framework as descried in the `Tags' section. In addition, we apply entropy regularization on $s^s$ to encourage the model to output confident predictions on unlabeled data. The overall loss is: 
$$ 
\mathcal{L}_I= \mathcal{L}_{\text{Tags}}(\phi, \hat{q}) + H(s^s) = -\sum_{c\in C}  \hat{q}(c) \log \phi(c) -\sum_{r\in R',c\in C} s^s(r,c)\log s^s(r,c),
$$ 
where $\hat{q}(c) = \delta(\phi(c) > \tau)$ and  $\delta(\cdot)$ is the delta function. 
As pointed out in \cite{tang2017multiple,Zeng_2019_ICCV,RadosavovicDGGH18}, self-ensembling is helpful when utilizing unlabeled data. We thus stack multiple ROI-classification and regression layers. Pseudo ground-truth will be computed from the ROI-classification logits of one layer to supervise another one. For inference, the average prediction is adopted.

\subsection{Proposal Refinement}
\label{sec:roi-refine}
Given strong labels, it's a standard technique~\cite{fastrcnn,ren16faster} to reject most false positive proposals and re-balance the training batch using the ground-truth boxes. However, proposal refinement using partial labels has not been studied before. 
Specifically, we keep a specific positive and negative proposal ratio  in each mini-batch. Positive proposals  satisfy two requirements: (1) one of the  ground-truth points or scribbles should be contained in each positive ROI; (2) all the selected positive ROIs together need to cover all the annotations. Negative proposals from the ROIs contain no labels. When generating a training batch we sample according to a pre-defined ratio. This practice dramatically decreases the number of proposals and thus  simplifies subsequent optimization. We refer to the proposal set after sampling and re-balancing using $R'$, as shown in Fig.~\ref{fig:archi} left.

%% file: sections/label.tex
\section{Partial Labels Simulation}
\label{sec:label}

Partial labels (\eg, points and scribbles) are much easier and natural to annotate than bounding boxes. They also provide much stronger localization information compared to tags. However, these types of annotations are either incomplete (\eg, part of the VOC images are labeled with points~\cite{train-click,extreme-click}) or missing (\eg, no partial labels have been annotated for COCO or LVIS) for object detection.

As a proof-of-concept for the proposed framework, we therefore develop an  approach to synthesize partial labels when ground-truth instance masks are available. It is our  goal  to mimic practical human labeling behavior. We are aware that the quality of the generated labels is sub-optimal. Yet these labels provide a surrogate to test and demonstrate the effectiveness of UFO$^2$. In this work, we generate the semantic partial labels for every object in the scene, and leave  manual collection of labels  to future work. 

\begin{figure}[t]
\centering
\includegraphics[width=\columnwidth]{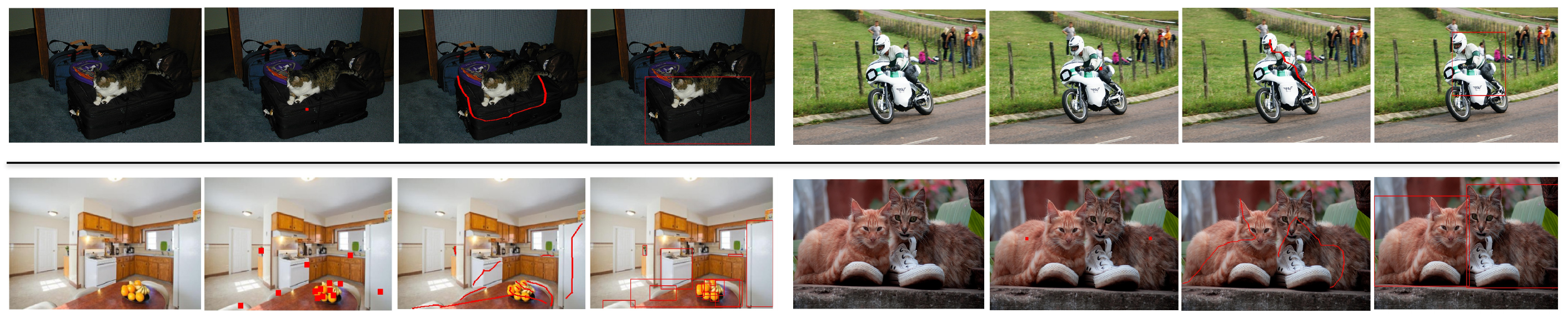} 
\caption{Top row: labels for single instance (suitcase and person). Bottom row: labels for all the objects (see appendix for more). }
\label{fig:label}
\end{figure}

\noindent\textbf{Points.} 
When annotating points, humans tend to click close to the center  of the objects~\cite{extreme-click}. However, different objects differ in shapes and poses.  Hence, their center usually does not coincide with the bounding box center. To mimic human behavior, we first apply a distance transform on each instance mask. The obtained intensity maps represent the distance between the points inside the body region and the closest boundary. This distance transform usually generates a `ridge' inside the object. We thus further normalize it and multiply with a Gaussian probability restricted to the bounding box. The final probabilistic maps are used to randomly sample one point as the annotation. 

\noindent\textbf{Scribbles.} 
Scribbles are harder to simulate since human annotators generate very diverse labels. Here we provide a  way to generate relatively simple scribbles. The obtained  labels likely don't perfectly mimic human annotations, yet they  serve as a proof-of-concept to show the effectiveness of the proposed framework. Given the instance mask, we first compute the topological skeleton, \ie, a connected graph, using OpenCV's~\cite{opencv_library} skeleton function. Using this graph, we start from a random point and seek a long path by extending in both directions. At intersections we randomly choose. We post-process the paths to avoid that their ends are close to the boundary. This latter constraint is inspired by the observation that humans usually don't draw scribbles close to the boundary. 

Representative generated labels are visualized in Fig.~\ref{fig:label}, where the top row shows examples for a single object (\ie, suitcase and rider) and the bottom row shows the labels for all objects in the scene. We observe the partial labels to be correctly located within each object. They also exhibit great diversity in terms of location and length across different instances. 

%% file: sections/experiments.tex
\section{Experiments}
We assess the proposed framework subsequently after detailing dataset, evaluation metrics and implementation.

\noindent\textbf{Dataset \& Evaluation Metrics.} 
We conduct experiments on COCO~\cite{mscoco} -- the most popular dataset for object  detection. Standard metrics are reported including AP (averaged over IoU thresholds) and AP-50 (IoU threshold at 50\%). We use several COCO splits in this paper:
(1) \texttt{COCO-80}: COCO 2014 train set of ~80K images.
(2) \texttt{COCO-35} (a.k.a.\ \texttt{valminusminival}): a 35K subset of  COCO 2017 train set.
(3) \texttt{COCO-115}: COCO 2017 train set, equals  union of  \texttt{COCO-80} and  \texttt{COCO-35}.
(4) \texttt{COCO-val}: COCO 2014 val set of ~40K images.
(5) \texttt{minival}: COCO 2017 val set of 5K images.
(6) \texttt{Un-120}: COCO unlabeled set of 120k images.

\noindent\textbf{Implementation Details.} For a fair comparison to prior work with different forms of a single supervision, we use the most common VGG-16 and ResNet-50 backbones. SGD is used for optimization.  After proposal refinement, we keep 1024 ROIs for points and scribbles and 512 for boxes as those have the most localization information and thus a reduced  need for abundant ROIs.

\subsection{Evaluation of Single Labels}
\label{exp:from-0}

\begin{table}[t]
\centering
\begin{tabular}{ c| c | c | c c}
\specialrule{.15em}{.05em}{.05em} 
Methods & Test-scale & Label & AP & AP-50\\
\hline
PCL~\cite{tang2018pcl} & multi & tags  & 8.5 & 19.4 \\
C-MIDN~\cite{Gao_2019_ICCV} & multi  & tags  & 9.6 & 21.4 \\
WSOD2~\cite{Zeng_2019_ICCV} & multi  & tags  & 10.8 & 22.7 \\
Ours & multi & tags  & \textbf{11.4} & \textbf{24.3}  \\
\hline
Ours & single & tags  &  \textbf{10.8}  &  \textbf{23.1}  \\
Ours & single & points  &  \textbf{12.4}  &   \textbf{27.0}  \\
Ours & single & scribbles &   \textbf{13.7}  &  \textbf{29.8} \\
\hline
Fast-RCNN~\cite{fastrcnn} & single  & boxes & 18.9 & 38.6 \\
Faster-RCNN~\cite{ren16faster} & single  & boxes & 21.2 & 41.5 \\
Ours & single & boxes &  \textbf{25.7} &  \textbf{46.3} \\
\specialrule{.15em}{.05em}{.05em} 
\end{tabular}
\caption{Training on \texttt{COCO-80} from scratch using a single form of annotation and testing on \texttt{COCO-val}. All  results are obtained with a VGG-16 backbone.}
\label{table:single_sup}
\end{table}

\subsubsection{Train From Scratch.} 
We fist study the scenario where each single supervision is used to train a model from scratch. A VGG-16 model is trained on \texttt{COCO-80} and evaluated on \texttt{COCO-val} for a fair comparison to both weakly-supervised (tags) and strongly-supervised (boxes) work.  The results are reported in Tab.~\ref{table:single_sup}. Following prior work, we report both single-scale and multi-scale testing results (`Test-scale' column in Tab.~\ref{table:single_sup}).

When using tags, our model performs comparable to prior work. We slightly increase AP and AP-50 by 0.6\% and 1.6\% (Tab.~\ref{table:single_sup} top block). When using strong labels, our method also outperforms Fast- and Faster-RCNN baselines (Tab.~\ref{table:single_sup} bottom block). We found improvements to be due to the strong teacher losses  introduced in Sec.~\ref{sec:box}. In addition, we also report the results of  partial labels (tags, points, and scribbles) in the center block of Tab.~\ref{table:single_sup}, where we observe a natural correlation of performance  with complexity of the labels. Note that the performance boost from tag to point (+1.6\% AP/3.9\%AP-50) is bigger than the boost from point to scribble (+1.3\% AP/2.8\%AP-50). Also, strong labels still result in the biggest performance boost: it is significantly larger than that of  partial labels. Hence, bounding boxes are  necessary for accurate performance. 

\begin{table}[t]
\centering
\begin{tabular}{c c c| c c | c  c c | c}
\specialrule{.15em}{.05em}{.05em} 
Train & Methods & backbone & Labels & AP & Extra & Labels & AP & $\Delta$\\
\hline
\texttt{COCO-35}  & ours & VGG-16 & tags & 4.9 & \texttt{COCO-80} & - & 5.3 & \textbf{8.2\%} \\
\texttt{COCO-115} & ours & VGG-16 & tags & 12.9 & \texttt{Un-120} & -  &  13.6 & \textbf{5.4\%} \\
\texttt{COCO-35} & ours & ResNet-50 & tags & 9.8 & \texttt{COCO-80} & - & 10.5 & \textbf{7.1\%} \\
\hline
\texttt{COCO-35} & ours & ResNet-50 & boxes & 29.1 & \texttt{COCO-80} & tags & 29.4 & \textbf{1.0\%} \\
\texttt{COCO-35} & ours & ResNet-50 & boxes & 29.1 & \texttt{COCO-80} & points  & 30.1  & \textbf{5.5\%} \\
\texttt{COCO-35} & ours & ResNet-50 & boxes & 29.1 & \texttt{COCO-80} & scribbles & 30.9 & \textbf{6.2\%} \\
\hline
\texttt{COCO-115} & ours & ResNet-50 & boxes & 32.7 & \texttt{Un-120} & -  & 33.9 & \textbf{3.7\%} \\
\specialrule{.15em}{.05em}{.05em} 
\end{tabular}
\caption{Fine-tuning to improve an existing model using each single supervision. Results are reported by testing on \texttt{minival}.}
\label{table:single_sup_ft}
\end{table}

\subsubsection{Improve Existing Models.} 
We now study the use of each label to boost a pre-trained object detector. Results are shown in Tab.~\ref{table:single_sup_ft}. This experimental setup   follows semi-supervised learning studies and mimics a common practical scenario: we want to apply a pre-trained object detector to new environments while keeping annotation cost low or while having weak labels readily available. Motivated by this scenario,  pre-trained models are only fine-tuned by integrating extra weaker labels (\eg, first train with boxes, and then fine-tune with points, scribbles, or unlabeled data). We study two cases  in Tab.~\ref{table:single_sup_ft}: (1) small scale: from \texttt{COCO-35} to \texttt{COCO-80} where the model sees more unlabeled data; (2) large scale: from \texttt{COCO-115} to \texttt{Un-120} where labeled and unlabeled data are of similar size. 

In Tab.~\ref{table:single_sup_ft} (top block), we show that unlabeled data can be utilized to improve the performance of a weakly-supervised model where the VGG-16 and ResNet-50 based model are improved by 8.2\% (relative) and 7.1\% (relative), respectively. In Tab.~\ref{table:single_sup_ft} (center), we further demonstrate that partial labels are effective for improving a strongly-supervised model. Similarly, the relative performance improvement from tag to point (+4.5\%$\Delta$) is bigger than the improvement from point to scribble (+0.7\%$\Delta$). In Tab.~\ref{table:single_sup_ft} (bottom), we  use unlabeled data for a strongly supervised ResNet-50 based model, where unlabeled data improves its performance by  1.2\% AP (3.7\% relative improvement).

\begin{figure*}[t]
\includegraphics[width=\textwidth]{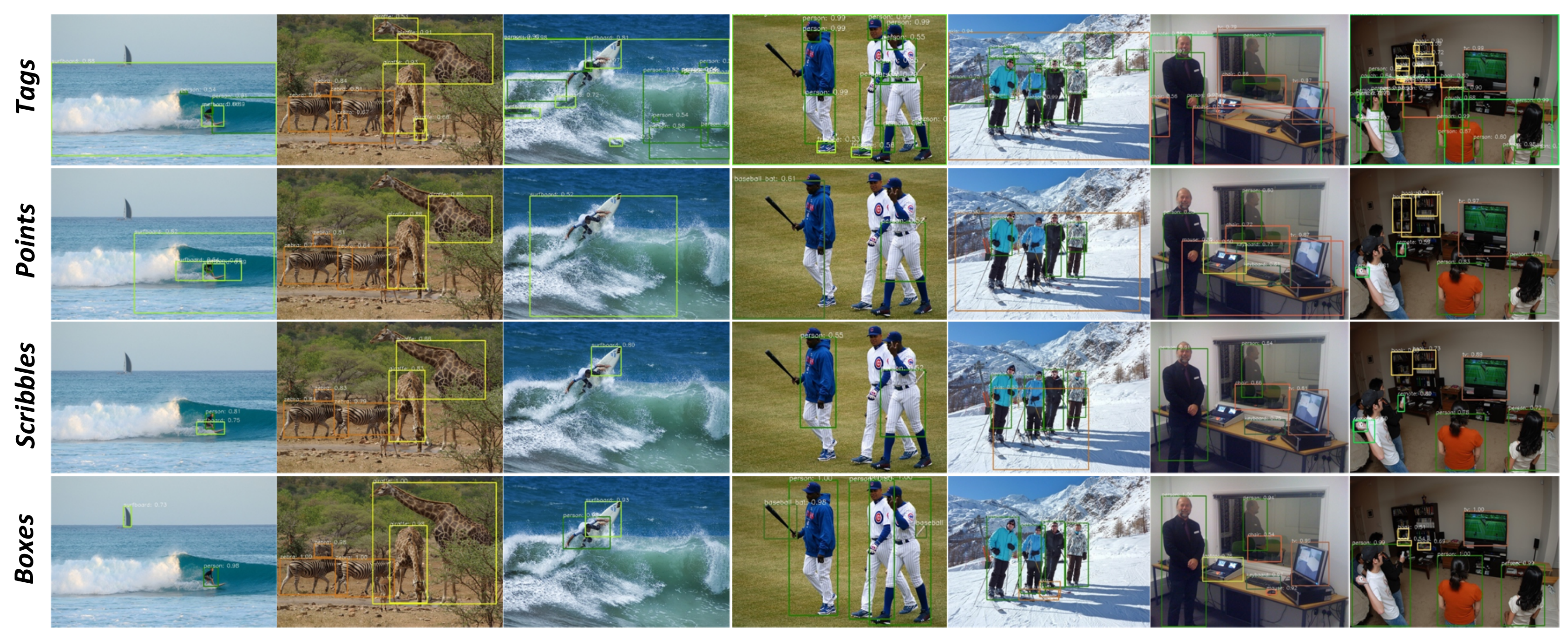} 
\caption{Qualitative comparison of models trained by using different labels. }
\label{fig:quali1}
\end{figure*}

\subsection{Qualitative Results}
\label{exp:quali}
Qualitative comparisons of the same model trained using different forms of supervision are shown in Fig.~\ref{fig:quali1}. From top to bottom we show predicted boxes and their confidence score when using tags,  points, scribbles, and boxes. We observe stronger labels to help the model  reject false positive predictions (\eg, the noisy small boxes on the sea and around the human head, the cars and books in the background), and also localize better  true positive predictions (\eg, the giraffe, surfing man, and each person in a crowd). More results and some failure modes are provided in the Appendix. 

\subsection{Ablation Study}
Next we study the effectiveness of each proposed module in the UFO$^2$ framework. 

\noindent{\textbf{Do strong teacher losses help?}} In Sec.~\ref{sec:box}, we introduce three extra losses for a balanced teacher student model when using boxes. Theses losses provide two advantages: (1) the model trained with strong labels  improves as reported in Tab.~\ref{table:teacher-ablation}.  These results are obtained using the same setting as those given in  Tab.~\ref{table:single_sup}. In the table, $\mathcal{L}_C+\mathcal{L}_R$ represents the vanilla version, \ie,  only ROI classification and regression heads are trained following supervised methods~\cite{fastrcnn,ren16faster}. We then  add each teacher loss  and illustrate that each one of them is beneficial. The best number is achieved when all three are combined. (2) strong losses  help omni-supervised learning. Without those losses, using tags, points, and scribbles will hurt the performance of a strongly pre-trained model by -5.6\%, -5.2\%, and -6.3\% compared to the performance improvement gained in Tab.~\ref{table:single_sup_ft} (middle). 

\begin{table}[t]
\centering
\begin{tabular}{c | c | c | c|  c | c }
\specialrule{.15em}{.05em}{.05em} 
Loss & $\mathcal{L}_C+\mathcal{L}_R$ & $+\mathcal{L}_{T1}$ & $+\mathcal{L}_{T2}$ & $+\mathcal{L}_{T1}+\mathcal{L}_{T2}$ & $+\mathcal{L}_{T1}+\mathcal{L}_{T2}+\mathcal{L}_{\text{Tags}}$ \\
\hline
AP   & 22.6 & 24.8 & 25.1 & 25.5 & 25.7 \\
AP-50 & 42.4 & 44.1 & 45.0 & 46.0  & 46.3 \\
\specialrule{.15em}{.05em}{.05em} 
\end{tabular}
\caption{Ablation study for the three strong teacher losses. }
\label{table:teacher-ablation}

\begin{tabular}{c | c |c c|  c c  }
\specialrule{.15em}{.05em}{.05em} 
Methods & Tags & Ref$_P$ & Ref$_S$ & Ref$_P$ + Con$_P$ & Ref$_S$  + Con$_S$  \\
\hline
AP   & 10.8 & 11.1 & 11.6 & 12.4 & 13.7 \\
AP-50 & 23.1 & 24.2 & 25.1 & 27.0 & 29.8 \\
\specialrule{.15em}{.05em}{.05em} 
\end{tabular}
\caption{Ablation study for localization constraints (Con$_P$ and Con$_S$ for points and scribbles) and proposal refinement (Ref$_P$, Ref$_S$). }
\label{table:partial-ablation}
\end{table}

\noindent{\textbf{Does proposal refinement help?}}
We evaluate the localization constraints proposed in Sec.~\ref{sec:ps} `Points \& Scribbles'  and the proposal refinement module (Sec.~\ref{sec:roi-refine}) following the settings of Tab.~\ref{table:single_sup}. The results are reported in  Tab.~\ref{table:partial-ablation}. Both proposed modules  improve the final performance. The localization constraints play a more important role than proposal refinement. This is reasonable as the localization constraints also consider the semantic information of partial labels. 

\subsection{Omni-supervised Learning}
\label{exp:pizza}

Given a fixed annotation budget, we can either choose to annotate more data with cheaper labels or less data with strong labels. With the proposed unified model, we empirically study and compare several annotation policies, and we provide a new insight regarding a suitable strategy.

\noindent\textbf{Annotation Time Estimation.} 
We use the labeling time as the annotation cost and ignore  other factors in this work. We  approximate the annotation time of each supervision relying on the annotation and dataset statistics reported in the literature~\cite{mscoco,bearman_point}. 

\begin{itemize}[noitemsep,topsep=0pt]
\item \textbf{Tags:} Collecting image-level class labels takes 1 second per category according to~\cite{bearman_point}. Thus, the expected annotation time on COCO is 80 sec/img.
\item \textbf{Points:} COCO~\cite{mscoco} contains 3.5 categories and 7.7 instances per image on average. Similarly to above, it takes 1 second to annotate every non-exist classes, for $80-3.5=76.5$ seconds in total.
\cite{bearman_point} reports that annotators take a median of 2.4 seconds to click on the first instance of a class, and 0.9 seconds for every additional instances. Thus the total labeling time is $76.5+3.5\times 2.4 + (7.7-3.5)\times 0.9=88.7$ sec/img. Note that point supervision is only 1.1 times more expensive than tags which is very efficient.
\item \textbf{Scribbles:} For each existing class, drawing a free-form scribble takes 10.9 seconds on average~\cite{LinDJHS16ScribbleSup,bearman_point}. Hence, the total time is $76.5+7.7\times 10.9=160.4$ sec/img. This number is roughly twice the time of labeling tags or points.
\item \textbf{Boxes:} It took 35s for one high quality box according to~\cite{HJL_AAAI12}. Hence, the total annotation time is $76.5+7.7\times 35=346$ sec/img. 
\end{itemize}
Given above approximations, we roughly know that annotating 1 image with bounding boxes takes as much time as annotating 4.33/3.9/2.16 images with tags/points/scribbles.

\noindent\textbf{Budget-aware Omni-supervised Detection.} 
We wonder: \emph{what annotation policy  maximizes  performance given a  budget?} Let's assume the total budget is fixed, \eg., 800,000 seconds. We empirically study several policies as listed below:
(1) \texttt{\textbf{MOST}}: we aim to maximize the number of images thus the entire budget is used to acquire tag annotation;
(2) \texttt{\textbf{STRONG}}: all the budget is  used to annotate bounding boxes, which is widely-adopted in practice;
(3) \texttt{\textbf{EQUAL}}: use one quarter of the budget for each label;
(4) \texttt{\textbf{EQUAL-NUM}}: same amount labeled for each.

Via above labeling time analysis and single-label experiments, we find that annotating points is a good choice among partial labels: roughly as efficient as tags but leads to better results; only half the price of scribbles but performs comparably well. 
We thus also study  scenarios with different combinations of boxes and points: 
(1) \texttt{\textbf{80\%B}}: 80\%  budget on boxes; 20\% on points;
(2) \texttt{\textbf{50\%B}}: 50\%  budget on boxes; 50\% on points;
(3) \texttt\textbf{{20\%B}}: 20\%  budget on boxes; 80\% on points.

As reported in Tab.~\ref{table:omni}, for the fixed budget of 800,000s, we first `annotate' (sample from \texttt{COCO-35}) 10,000 images with tags for the \texttt{MOST} policy. The other settings will only annotate less images. Annotations are then sampled from those 10,000 images. For example, \texttt{STRONG} will annotate 2,312 images with boxes and the rest remains unlabeled, which will also be utilized in our method given in Sec.~\ref{sec:un}. Therefore, training will use the same 10,000 images, albeit different policies make use of  different labels.  A VGG-16 based model is trained as described above and evaluated on \texttt{minival}. 

\begin{table}[t]
\centering
\begin{tabular}{ c | c c | c }
\specialrule{.15em}{.05em}{.05em} 
Policy & Image Amount & Labels & AP\\
\hline
\texttt{MOST} & $10000$  & T  & 3.0 $\pm$ 0.57 \\
\texttt{STRONG} & $2312+7688$ & B+U  & 13.97 $\pm$ 0.98 \\
\texttt{EQUAL} & $2500+2255+1250+578+5417$  & T+P+S+B+U & 5.87 $\pm$ 0.70\\
\texttt{EQUAL-NUM} & $1185 \times 4 + 5260$  & T+P+S+B+U & 9.43 $\pm$ 0.68 \\
\hline
\texttt{80\%B} & $1804+1850+6346$ & P+B+U & \textbf{ 14.11 $\pm$ 1.01 } \\
\texttt{50\%B} & $4510+1156+4334$ & P+B+U & 11.13 $\pm$ 1.12 \\
\texttt{20\%B} & $7215+462+2323$ & P+B+U & 4.47 $\pm$ 0.75 \\
\specialrule{.15em}{.05em}{.05em} 
\end{tabular}
\caption{Budget-aware Omni-supervised Detection (T: tags, P: points, S: scribbles, B: boxes, U: unlabeled). Mean and standard deviation of Average Precision (AP) are reported over three runs.}
\label{table:omni}
\end{table}

We observe: (1) strong labels are still very important and the policy \texttt{STRONG} outperforms other popular polices by a great margin (Tab.~\ref{table:omni} top half). (2) It's not necessary to annotate every image with boxes to achieve competitive results.  \texttt{80\%B} is  slightly better than  \texttt{STRONG} and \texttt{50\%B} also performs better than \texttt{EQUAL-NUM}. This result suggests that spending a certain amount of cost annotating more images with points is a better annotation strategy than the  commonly adopted bounding box annotation (\texttt{STRONG}). 

%% file: sections/conclusion.tex
\section{Conclusions}
We present UFO$^2$, a novel unified framework  for omni-supervised object detection. It  handles strong labels, several forms of partial annotations (tags, points, and scribbles), and unlabeled data simultaneously.  UFO$^2$ is able to utilize each label effectively, permitting to study budget-aware omni-supervised object detection. We also assess a promising annotation policy. 

%% file: sections/supp.tex
\noindent\textbf{\Large{Appendix}}

\appendix

\noindent{}In this document, we provide: 
\begin{itemize}
\item An extensive study on the \texttt{LVIS}~\cite{lvis} dataset where UFO$^2$ learns to detect more than 1k different objects without bounding box annotation (Sec.~\ref{exp:extension}), which demonstrates the  generalizability and applicability of UFO$^2$
\item Details regarding the annotation policies  mentioned in Sec.~\ref{exp:pizza} `Budget-aware Omni-supervised Detection' of the main paper
\item Additional visualizations of the simulated partial labels
\item Additional qualitative results on \texttt{COCO} (complementary to main paper Sec.~\ref{exp:quali})
\end{itemize}

\input{sections/extension.tex}

\section{Annotation Policies}
To study budget-aware omni-supervised object detection, we defined the following policies: \texttt{80\%B}, \texttt{50\%B}, \texttt{20\%B} motivated by the following findings: 
(1) among the three partial labels (tags, points, and scribbles), labeling of points (88.7 s/img;  see Sec.~\ref{exp:pizza} in the main paper) is roughly as efficient as annotating tags (80 s/img), both of which require half the time/cost of scribbles (160.4 s/img); 
(2) using points  achieves a consistent performance boost compared to using tags (12.4 over 10.8 AP in Tab.~\ref{table:single_sup}; 30.1 over 29.4 AP in Tab.~\ref{table:single_sup_ft});  
(3)  using scribbles is just slightly better than using points (13.7 over 12.4 AP in Tab.~\ref{table:single_sup}; 30.9 over 30.1 AP in Tab.~\ref{table:single_sup_ft}) but twice as expensive to annotate; and 
(4) strong supervision (boxes) is still necessary to achieve good results (strongly supervised models are significantly better than others in Tab.~\ref{table:single_sup} and  Tab.~\ref{table:single_sup_ft}). 

Therefore, we choose to combine points and boxes as a new annotation policy which we found to work well under the fixed-budget setting as shown in Tab.~\ref{table:omni}: \texttt{80\%B} is slightly better than \texttt{STRONG} and \texttt{50\%B} also performs better than \texttt{EQUAL-NUM}. These results suggest that spending some amount of cost to annotate more images with points is a better annotation strategy than the  commonly-adopted bounding box only annotation (\texttt{STRONG}). Meanwhile, the optimal annotation policy remains an open question and better policies may exist if more accurate scribbles are collected or advanced algorithms are developed to utilize partial labels.

\section{Additional Visualization of Partial Labels }
\begin{figure}[t]
\centering
\includegraphics[width=\textwidth]{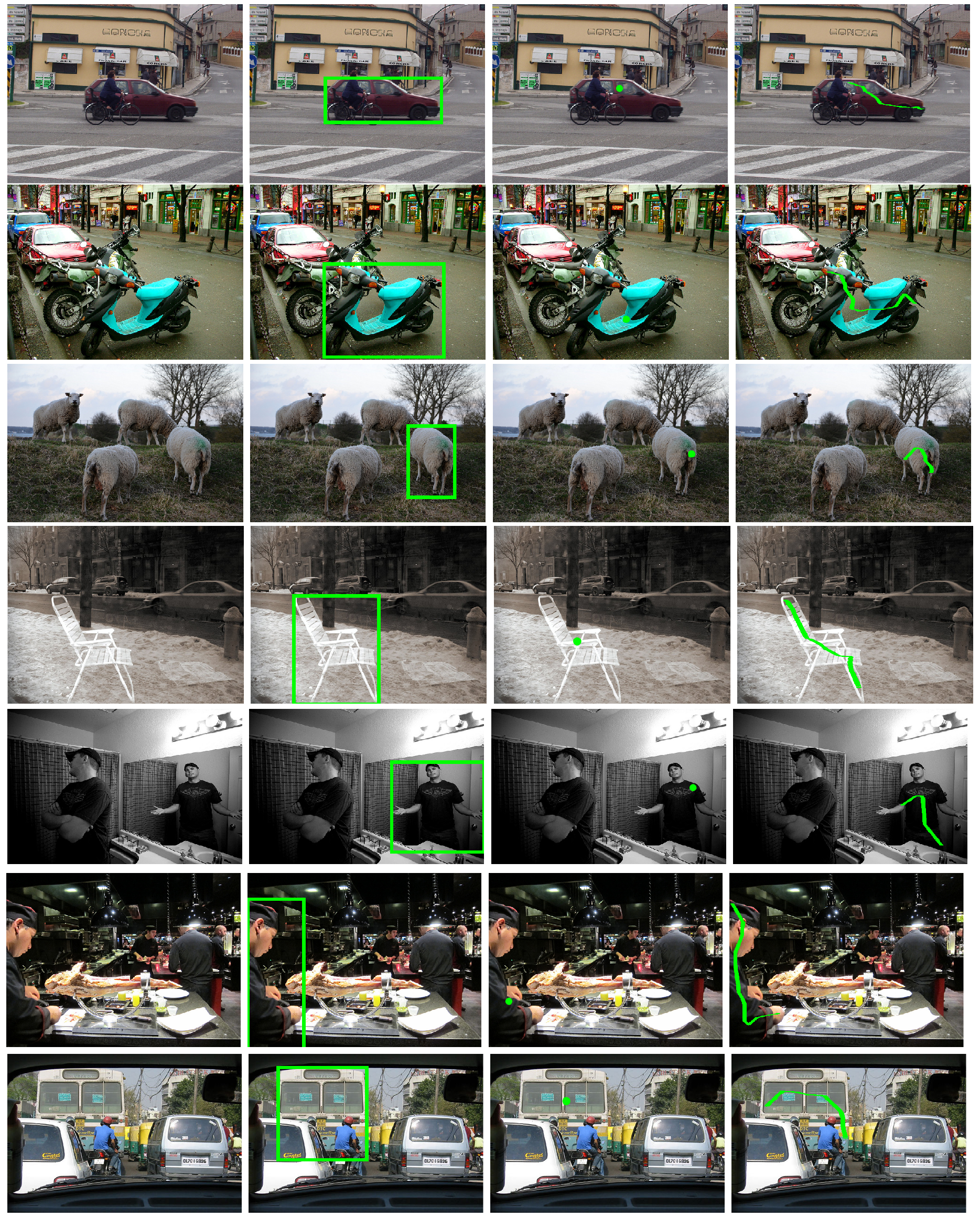}
\caption{Additional visualization of the ground-truth boxes and the simulated partial labels.}
\label{fig:quali_more}
\end{figure}

\begin{figure}[t]
\centering
\includegraphics[width=0.96\textwidth]{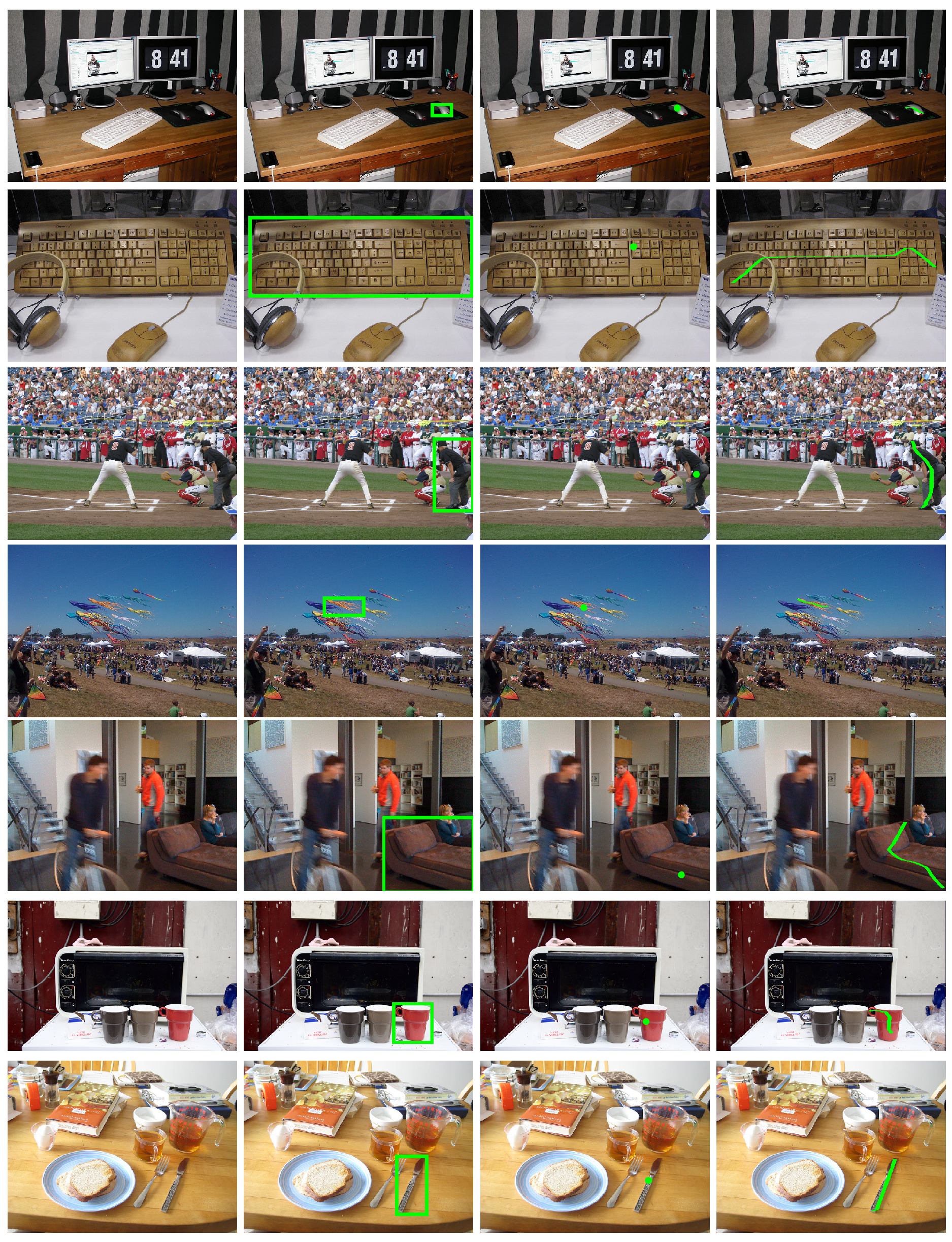}
\caption{Additional visualization of the ground-truth boxes and the simulated partial labels.}
\label{fig:quali_more_1}
\end{figure}

\begin{figure}[t]
\centering
\includegraphics[width=0.95\textwidth]{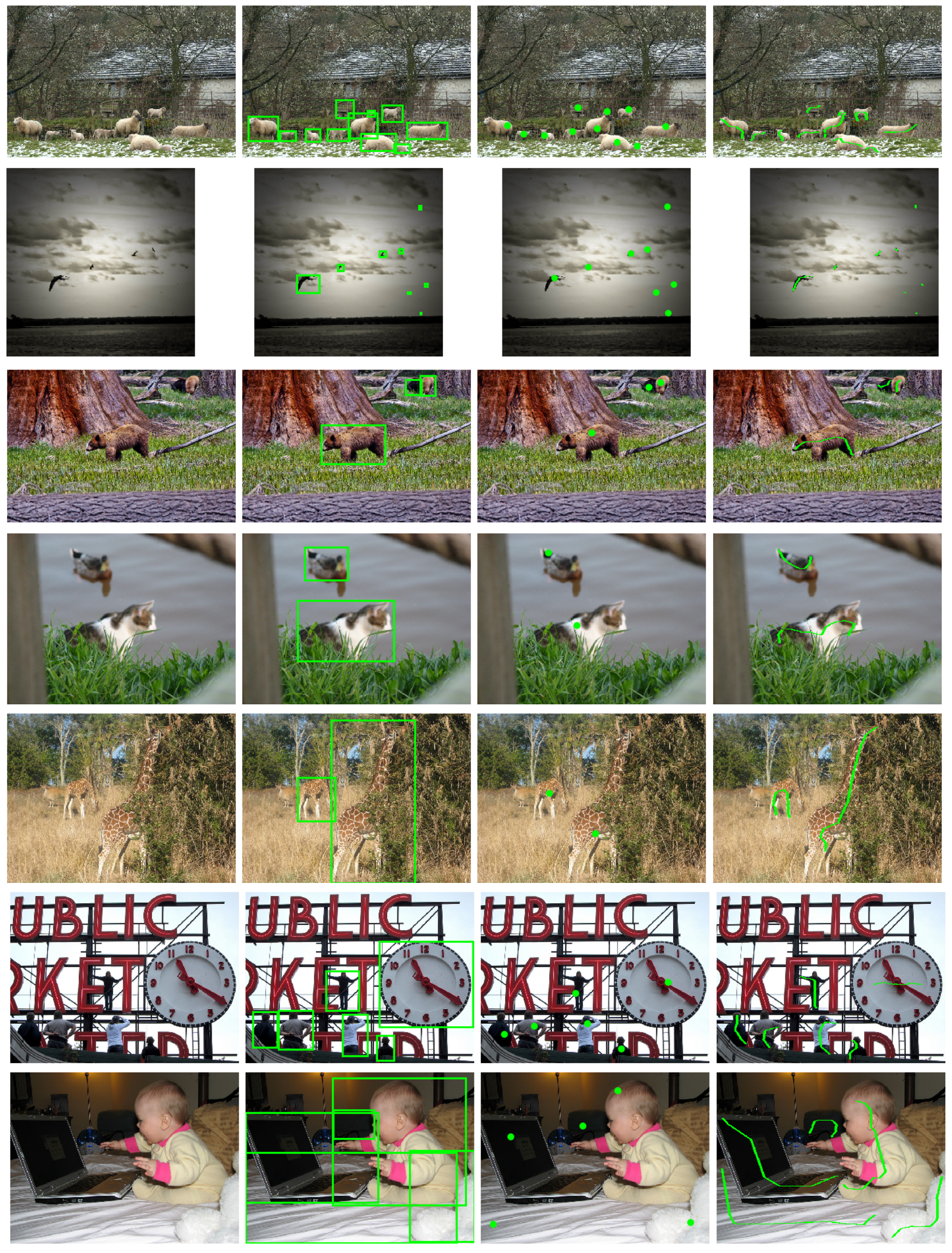}
\caption{Additional visualization of the ground-truth boxes and the simulated partial labels.}
\label{fig:quali_more2}
\end{figure}

\begin{figure}[t]
\centering
\includegraphics[width=0.98\textwidth]{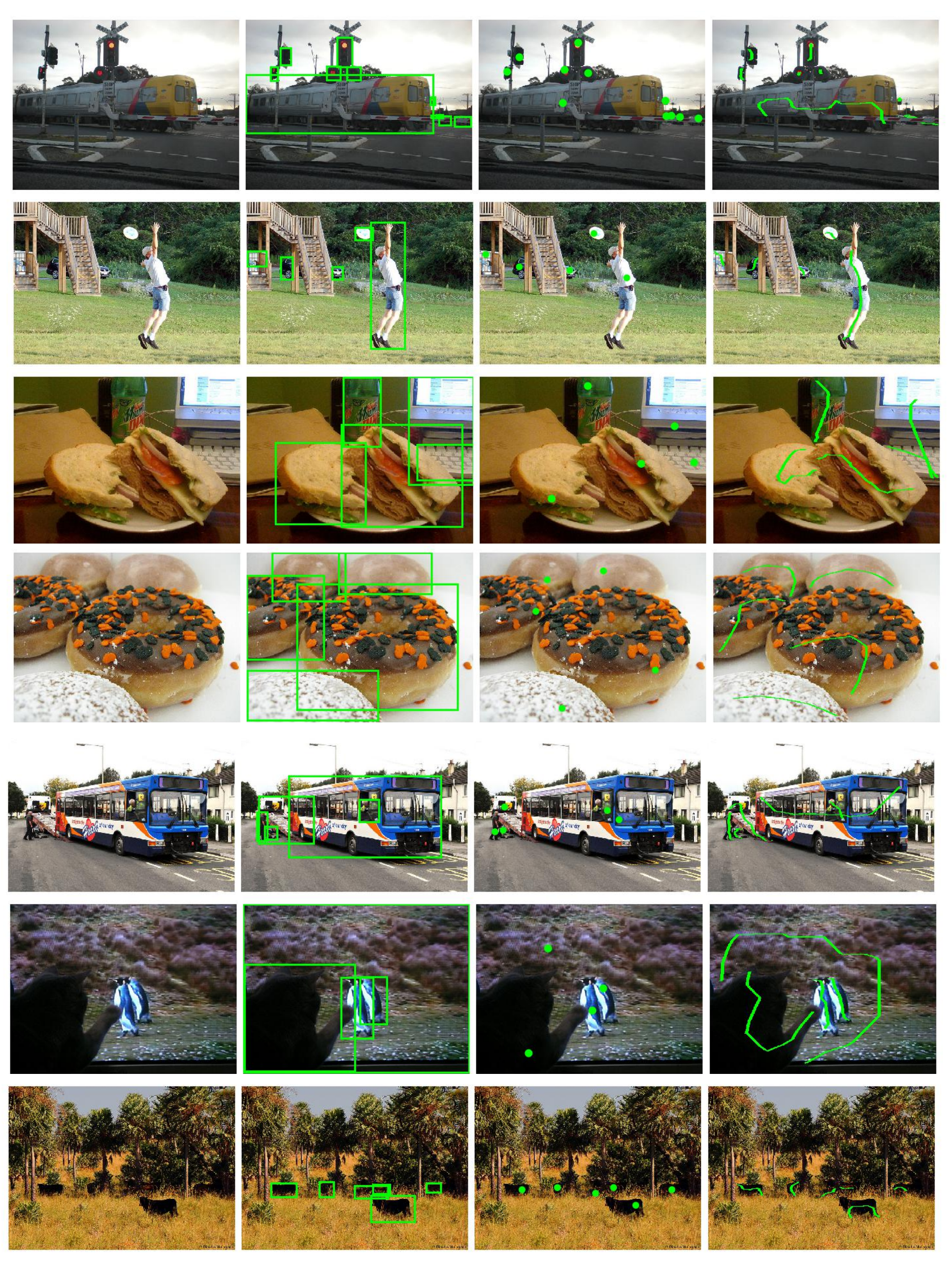}
\caption{Additional visualization of the ground-truth boxes and the simulated partial labels.}
\label{fig:quali_more2_1}
\end{figure}

We show additional results together with the ground-truth bounding boxes in Figs.~\ref{fig:quali_more}--\ref{fig:quali_more2_1}. Fig.~\ref{fig:quali_more} and Fig.~\ref{fig:quali_more_1} show labels for single objects (\eg, car, motor, sheep, chair, person, and bus) and Fig.~\ref{fig:quali_more2} and Fig.~\ref{fig:quali_more2_1} visualize labels for all the instances in the images. 

We observe: (1) both points and scribbles are correctly located within the objects; (2) points are mainly located  around the center area of the objects and with a certain amount of randomness, which aligns with our goal  to mimic human labeling behavior as discussed in Sec.~\ref{sec:label} of the main paper; 
(3) the generated scribbles are relatively simple yet effective in capturing the rough shape of the objects. Also, they exhibit a reasonable diversity. These partial labels serve as a proof-of-concept to show the effectiveness of the proposed UFO$^2$ framework. 

\section{Additional Qualitative Results}
\begin{figure}[t]
\centering
\includegraphics[width=\textwidth]{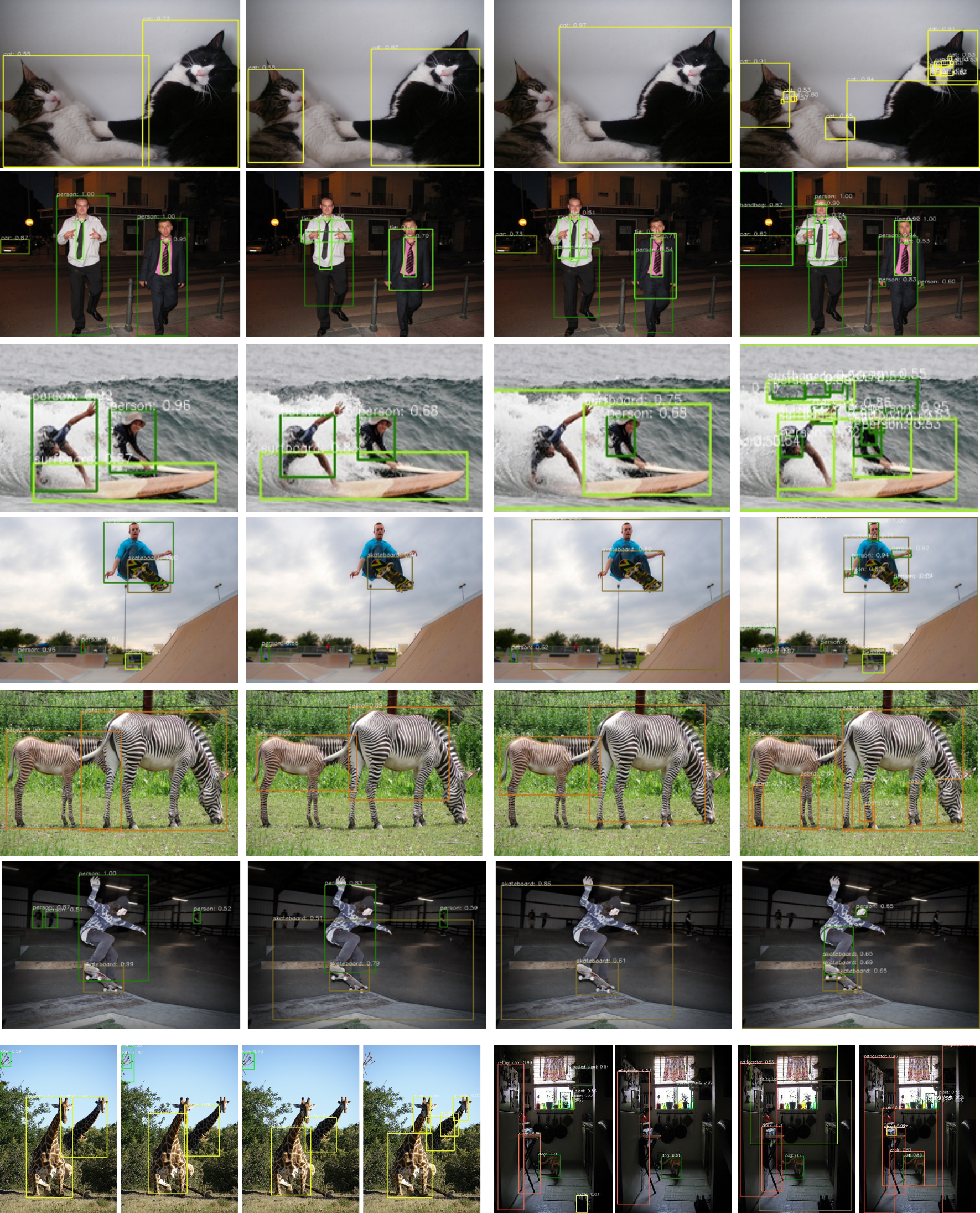}
\caption{Additional qualitative comparison of models trained with different labels on \texttt{COCO} (left to right: boxes, scribbles, points, tags).}
\label{fig:quali_more3}
\end{figure}

\begin{figure}[t]
\centering
\includegraphics[width=0.99\textwidth]{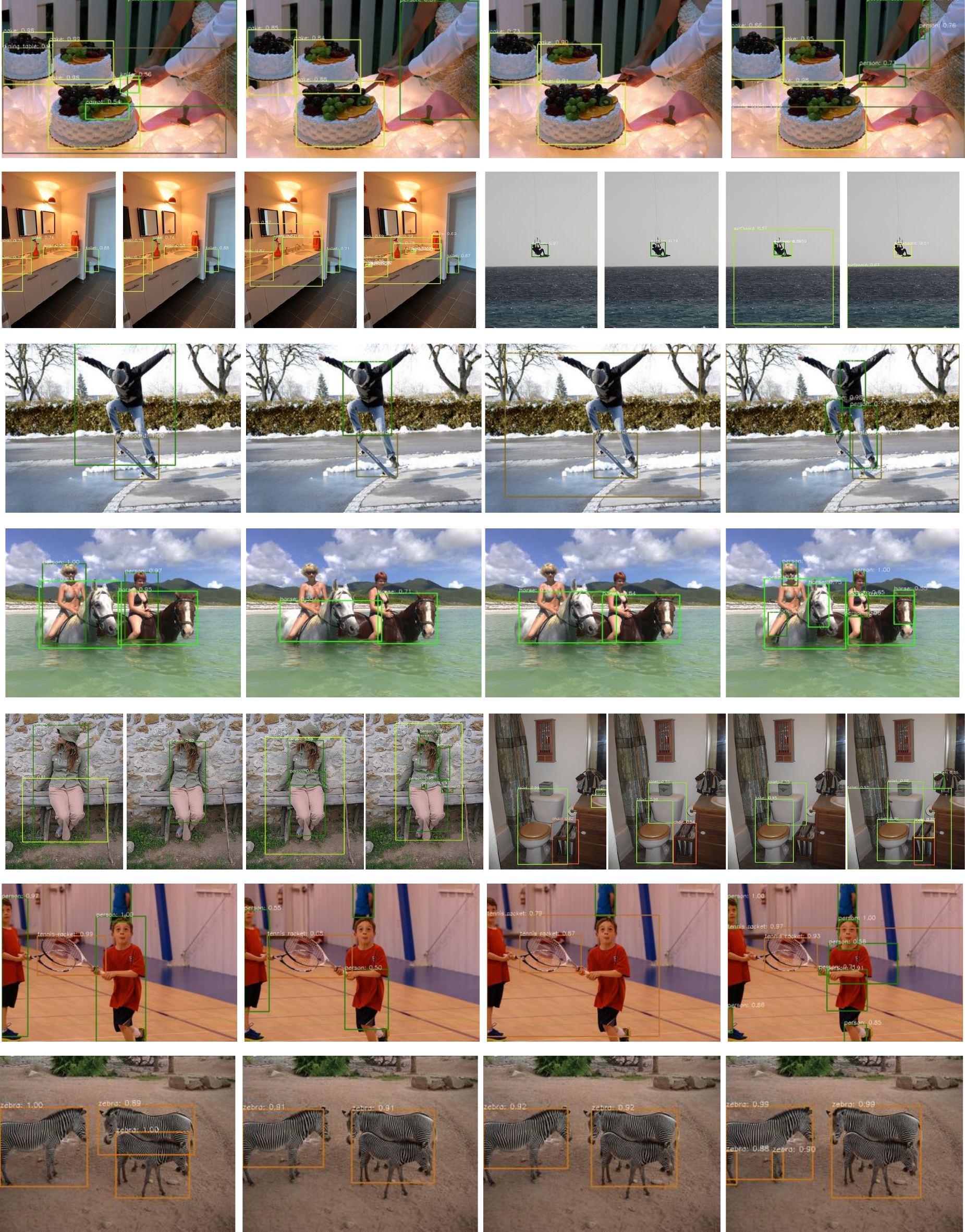}
\caption{Additional qualitative comparison of models trained with different labels on \texttt{COCO} (left to right: boxes, scribbles, points, tags).}
\label{fig:quali_more3_1}
\end{figure}

In Fig.~\ref{fig:quali_more3} and Fig.~\ref{fig:quali_more3_1} we show additional qualitative results. We compare the same VGG-16 based model trained on \texttt{COCO-80} with different forms of supervision. 
From left to right we show predicted boxes and their confidence scores when using boxes, scribbles, points, and tags. Similar to the results in Sec.~\ref{exp:quali} of the main paper, we find that stronger labels  better reduce false positive predictions and better localize true positive predictions.

%% file: sections/extension.tex
\section{Extensions: Learning to Detect Everything}
\label{exp:extension}
In this section we show that without any architecture changes UFO$^2$ can be generalized to detect any objects given image-level tags. Hence we follow~\cite{hu2018learning} and refer to this setting as `learning to detect everything.' 

Specifically, we aim to detect objects from the \texttt{LVIS}~\cite{lvis} dataset: it contains 1239 categories and we only use tags as annotation. We use \texttt{COCO*}\footnote[4]{\texttt{COCO*}: because  \texttt{LVIS} is a subset of \texttt{COCO-115}, we construct \texttt{COCO*} by taking \texttt{COCO-115} images excluded from \texttt{LVIS}.} with boxes and train UFO$^2$ on it first. Our model achieves 32.7\%AP and 52.3\%AP-50 on \texttt{minival}. We then jointly fine-tune this model using tags from \texttt{LVIS} and boxes from \texttt{COCO*}. The final model performs comparably on \texttt{minival} (31.6\%AP, 50.1\%AP-50) and also decent on \texttt{LVIS} validation of over 1k classes (3.5\%AP, 6.3\%AP-50 where a supervised model achieves 8.6\%AP, 14.8\%AP-50). To the best of our knowledge, no numbers have been reported on this dataset using weak labels. Our results are also not directly comparable to strongly supervised results~\cite{lvis} as we don't use the bounding box annotation on \texttt{LVIS}.

\begin{figure}[t]
\centering
\includegraphics[width=0.97\textwidth]{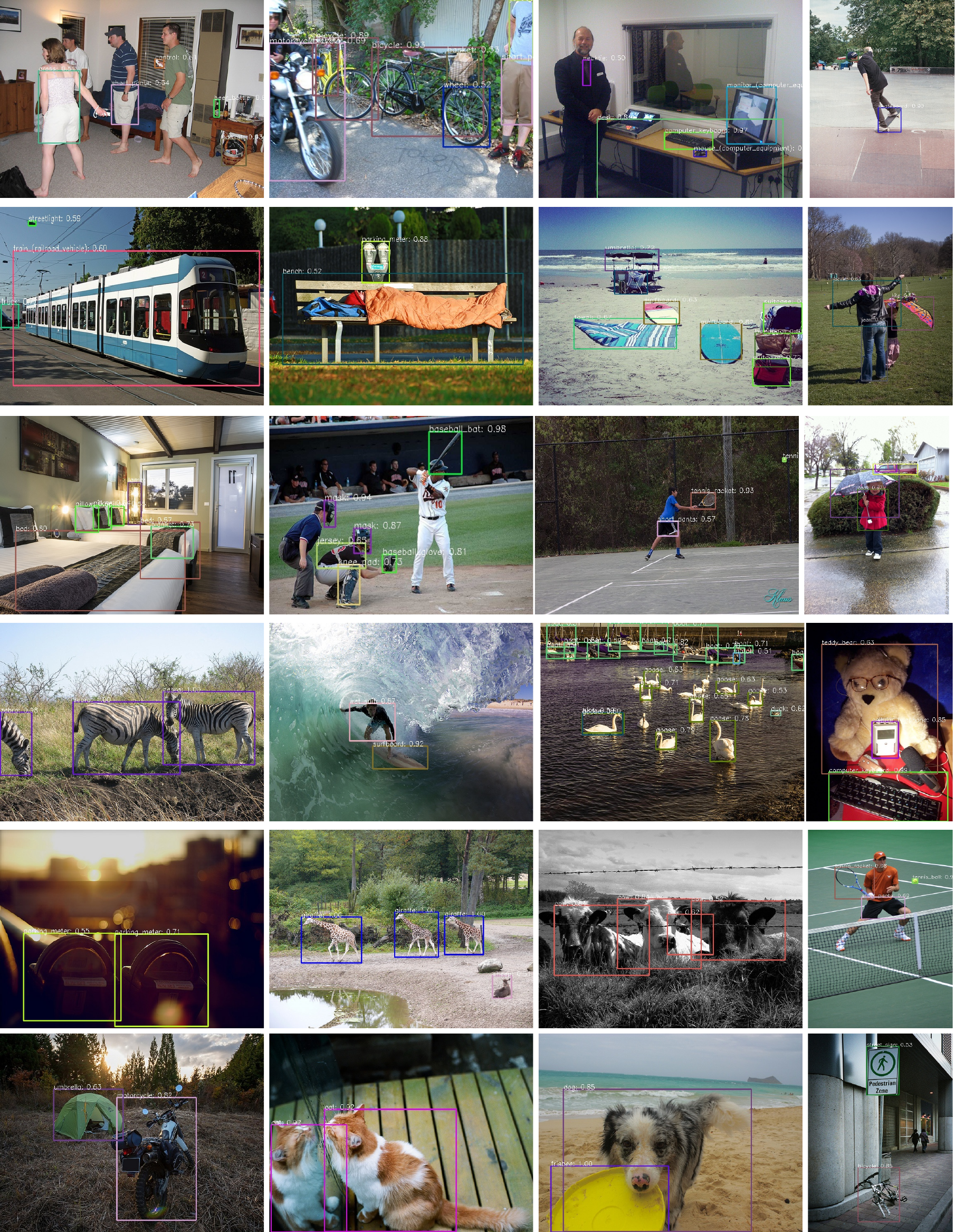}
\caption{Visualization of results on \texttt{LVIS} data.}
\label{fig:quali_more4}
\end{figure}

Qualitative results are shown in Fig.~\ref{fig:quali_more4}. We observe that UFO$^2$ is  able to detect objects accurately even though no bounding box supervision is used for the new classes (\eg, short pants, street light, parking meter, frisbee, \etc). Specifically, UFO$^2$ can (1) detect spatially adjacent or even clustered instances with great recall (\eg, goose, cow, zebra, giraffe); (2) recognize some obscure or hard objects (\eg, wet suite, short pants, knee pad); (3) localize different objects with tight and accurate bounding boxes.
Importantly, note that we don't need to change the architecture of UFO$^2$ at all to integrate both boxes and tags as supervision.

